\documentclass{article}

\usepackage{arxiv}

\usepackage{makecell}
\usepackage[utf8]{inputenc} 
\usepackage[T1]{fontenc}    
\usepackage{hyperref}       
\usepackage{url}            
\usepackage{booktabs}       
\usepackage{amsfonts}       
\usepackage{nicefrac}       
\usepackage{microtype}      
\usepackage{graphicx}
\usepackage{doi}
\usepackage[numbers]{natbib}
\usepackage{setspace}
\usepackage{amsmath}
\usepackage{amssymb}
\usepackage{eucal}
\usepackage{tcolorbox}
\tcbset{
  myquestionbox/.style={
    colback=gray!5,
    colframe=black,
    boxrule=0.5pt,
    arc=2pt,
    left=4pt,
    right=4pt,
    top=4pt,
    bottom=4pt,
    fonttitle=\bfseries,
    before skip=8pt,
    after skip=8pt
  }
}
\usepackage[export]{adjustbox}
\usepackage{multirow}
\usepackage{booktabs}  
\usepackage{rotating}
\usepackage{soul,xcolor}
\newcommand{\hlyellow}[1]{{\sethlcolor{yellow}\hl{#1}}}

\newcommand{\hlred}[1]{{\sethlcolor{red}\hl{#1}}}

\newcommand{\hlgreen}[1]{{\sethlcolor{green}\hl{#1}}}

\usepackage{algorithm}
\usepackage{algpseudocode}
\usepackage{listings}
\usepackage{xspace}

\def\tsc#1{\csdef{#1}{\textsc{\lowercase{#1}}\xspace}}
\tsc{WGM}
\tsc{QE}
\tsc{EP}
\tsc{PMS}
\tsc{BEC}
\tsc{DE}

\title{Knowledge Graph Fusion with Large Language Models for Accurate, Explainable Manufacturing Process Planning}

\date{} 					

\author{Danny Hoang\\
	School of Mechanical, Aerospace, and Manufacturing Engineering\\
	University of Connecticut\\
	Storrs, Connecticut, USA\\
	\And
David Gorsich \\
	The United States Army Combat Capabilities Development Command\\
	Ground Vehicle Systems Center\\
	Warren, Michigan, USA\\
	\AND
	Matthew P.~Castanier\\
The United States Army Combat Capabilities Development Command\\
	Ground Vehicle Systems Center\\
	Warren, Michigan, USA\\
	\And
	Farhad Imani \thanks{Corresponding author: farhad.imani@uconn.edu\\ DISTRIBUTION STATEMENT A. Approved for public release; distribution is unlimited. OPSEC9768}\\ School of Mechanical, Aerospace, and Manufacturing Engineering\\
	University of Connecticut\\
	Storrs, Connecticut, USA\\
}

\renewcommand{\headeright}{}
\renewcommand{\undertitle}{}
\renewcommand{\shorttitle}{KG-Fused LLMs for Explainable Process Planning}

\hypersetup{
pdftitle={Knowledge-Graph Fusion with Large Language Models for Accurate, Explainable Manufacturing Process Planning},
pdfauthor={Danny Hoang, David Gorsich, Matthew P. Castanier, Farhad Imani},
pdfkeywords={Large Language Model, Retrieval Augmented Generation, Knowledge Graph, Manufacturing Process Planning
}
}

\begin{document}
\maketitle
\begin{abstract}
Precision process planning in Computer Numerical Control (CNC) machining demands rapid, context-aware decisions on tool selection, feed-speed pairs, and multi-axis routing, placing immense cognitive and procedural burdens on engineers from design specification through final part inspection. Conventional rule-based computer-aided process planning and knowledge-engineering shells freeze domain know-how into static tables, which become limited when dealing with unseen topologies, novel material states, shifting cost-quality-sustainability weightings, or shop-floor constraints such as tool unavailability and energy caps. Large language models (LLMs) promise flexible, instruction-driven reasoning for tasks such as G-code synthesis to spindle-load queries, but they routinely hallucinate numeric values and provide no provenance. We present Augmented Retrieval Knowledge Network Enhanced Search \& Synthesis (ARKNESS), the end‑to‑end framework that fuses zero‑shot Knowledge Graph (KG) construction with retrieval‑augmented generation to deliver verifiable, numerically exact answers for CNC process planning. ARKNESS (1) automatically distills heterogeneous machining documents, handbooks, G-code annotations, and vendor datasheets into augmented triple, multi-relational graphs without manual labeling, and (2) couples any on-prem LLM with a retriever that injects the minimal, evidence-linked subgraph needed to answer a query. Benchmarked on 155 industry-curated questions spanning tool sizing, feed-speed optimization, and tolerance diagnostics, a lightweight 3B-parameter Llama-3 augmented by ARKNESS matches GPT-4o accuracy while achieving a +25 Percentage Point (pp) gain in multiple-choice accuracy, +22.4 pp in F1, and 8.1× ROUGE-L on open-ended responses. Additionally, by grounding reasoning in precise triples, ARKNESS enables smaller models to match or exceed the accuracy of much larger cloud models, reducing numeric hallucinations by 22 pp, and running fully on-premise for privacy-preserving, real-time inference on the shop floor.
\end{abstract}


\section{Introduction}

Modern production processes, such as precision machining, demands accuracy margins and tolerates virtually no numerical error, imposing substantial mental and operational loads on engineers from initial design specifications to the final part evaluation~\cite{wiessner2018thermal,hoang2024hierarchical,chen2024distributed}. In high-stakes sectors, such as aerospace and energy, dimensional tolerances routinely tighten to ± 5 $\mu$m, while surface-integrity constraints, white-layer thickness, and residual stress, must remain within narrowly defined thresholds~\cite{li2025sub}. To meet these criteria, machinists must simultaneously optimize a variety of intertwined process parameters, such as tool type, feed rate, spindle speed, and depth of cut, carefully trading off cycle time against surface finish, tool wear, and geometric fidelity~\cite{forinsights2023cnc,hoang2023edge,hoang2024brief}. An incorrect decimal drill size or an outdated cutting-speed chart therefore translates directly into scrap, rework or latent defects that may only become apparent after catastrophic failure~\cite{spanaki2025data}. Such unplanned downtime erodes an estimated 11\% of annual revenue for Fortune Global 500 manufacturers, roughly US \$1.5 trillion lost each year~\cite{siemens2022truecost}. 

Computer‑Aided-Process-Planning (CAPP) engines and the Computer-Aided Manufacturing (CAM) modules bundled within mainstream Computer-Aided Design (CAD) suites (e.g., Siemens NX, CATIA, and SolidWorks) form automation pipeline that ingests STEP‑compliant product models (AP203/242) and emits executable Numerical Control (NC) results~\cite{nwasuka2024computer}. A dedicated CAPP system executes the high‑level analysis loop, geometry interrogation, feature recognition, precedence, constraint resolution, raw‑stock/fixture/machine assignment, and operation‑graph synthesis, before compiling feeds, speeds, and tool‑change cycles into ISO6983 or STEP‑NC. Embedded CAM modules then refine the middle layers: selecting cutters from vendor libraries, applying parameter presets, generating and simulating toolpath with collision checks and material‑removal estimates, and invoking feature‑based machining or knowledge‑based‑engineering rules for common prismatic features. Collectively, these rule‑driven frameworks can compress planning lead time by 2–10×, yet they share a structural brittleness. Hard‑coded heuristics falter when faced with unseen topologies (e.g., self‑intersecting free‑form pockets), novel material states, shifting cost–quality–sustainability weightings, or shop-floor constraints, such as tool unavailability and energy caps. Moreover, neither tier natively ingests telemetry (e.g., cutting force, vibration, and thermal imagery) generated by machining cells, forcing human experts to intervene, reconcile documentation, and manually retune parameters, an increasingly untenable bottleneck as geometrical complexity and throughput requirements escalate.

\begin{figure}[t!]
    \centering
    \includegraphics[width=\textwidth]{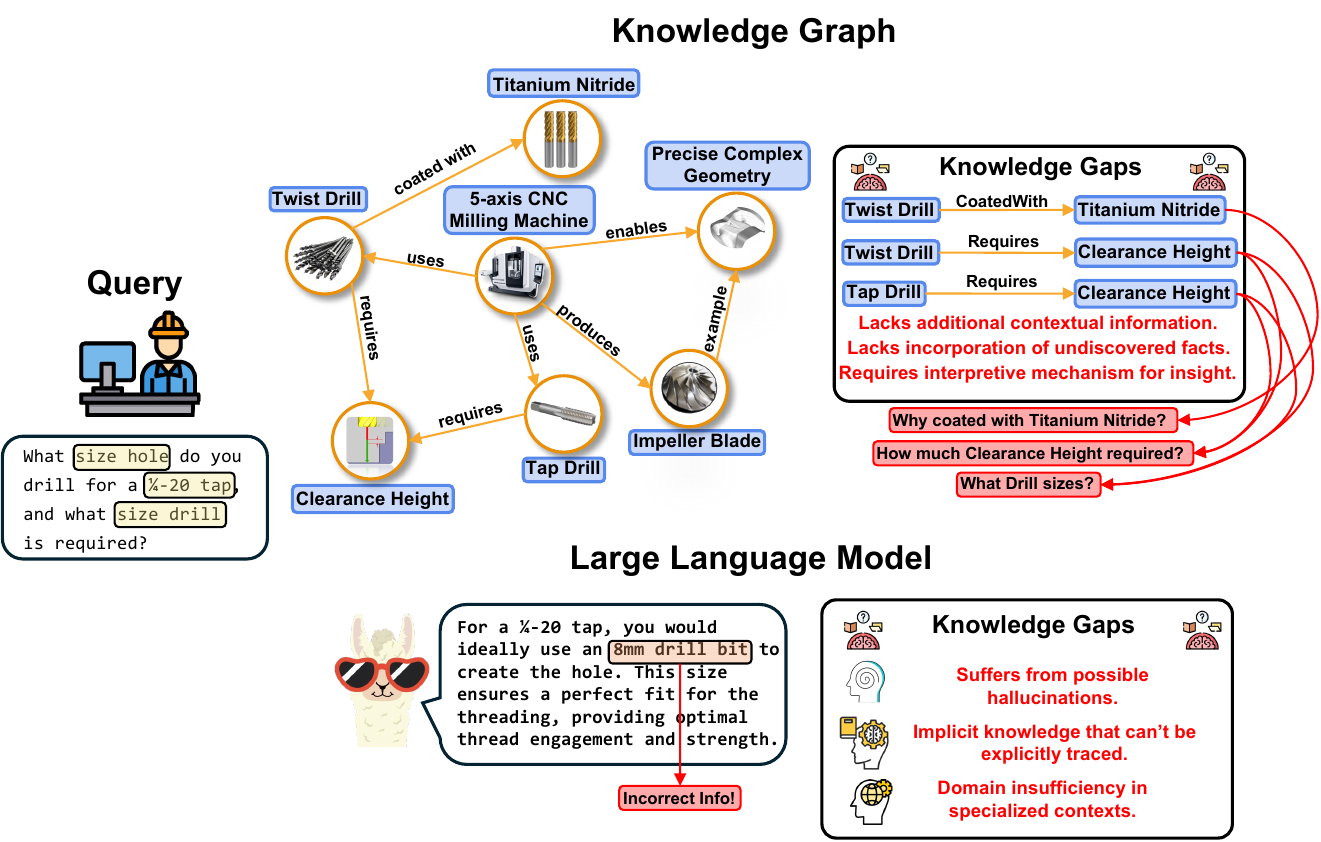}
     \vspace{-3mm}
    \caption{Illustration contrasting a knowledge graph (top) and a large language model (bottom) handling the same tap‑drill sizing query. In the knowledge graph section, technical entities (i.e., Twist Drill, Clearance Height, Tap Drill, Titanium Nitride, 5‑axis CNC Milling Machine, and Impeller Blade) are connected by explicit relations but can often be missing contextual links and undiscovered facts. The LLM section shows the model’s generated answer (“8 mm drill bit”), annotated to expose imprecision risks, implicit untraceable knowledge, and domain inefficiency in specialized machining.}
     \vspace{-3mm}
    \label{fig:why}
\end{figure}

Large Language Models (LLMs) have emerged as versatile, instruction‑following agents capable of synthesizing domain knowledge, reasoning under uncertainty, and generating executable results~\cite{raza2025industrial}. Recent studies demonstrate their ability to synthesize 3-axis drilling paths, to answer contextual queries about spindle loads and feed rates, and to troubleshoot controller alarms with human-level accuracy in descriptive tasks~\cite{vsket2024enhancing}. Šket et al. benchmarked GPT‑3.5 and GPT‑4 across independent G‑code generation, self‑interpretation, and error simplification; GPT‑4 rendered more correct toolpath yet remained limited to simple drilling operations and required extensive prompt steering~\cite{vsket2024enhancing}. To broaden functionality beyond code synthesis, ChatCNC was introduced, coupling multiple LLM agents with real‑time machine telemetry so that an operator can, for example, ask “What was the spindle load at 09:32?”; the system attained 93.3\% accuracy on complex production‑tracking queries~\cite{jeon2025chatcnc}. Nevertheless, both studies expose two systemic bottlenecks: first, dependence on precise user prompts or narrowly structured SQL retrieval, and second, fragile performance when required context is absent from the local database.

Domain‑specific fine‑tuning offers one mitigation strategy. Rosati et al. raised ROUGE‑1 scores of Llama‑3 from 0.164 to 0.314 using industrial user manuals~\cite{rosati2024improving}. CNCGPT fine‑tuned GPT‑3.5 Turbo on machine instructions and trouble‑shooting logs, boosting factual correctness from ROUGE‑L 0.296 to 0.692~\cite{kanimozhi2024explorative}. Yet intensive fine‑tuning presents new trade‑offs: (1) escalating computational cost as the query‑response inventory grows into the thousands; (2) rapid obsolescence that mandates frequent retraining; and (3) data‑sovereignty concerns when proprietary part files or production metrics traverse external cloud APIs. Hence, a complementary mechanism is required, one that supplies rich, verifiable context to an LLM without chronically retraining the model or exposing sensitive data.

Knowledge Graphs (KGs), on the other hand, provide a structured, semantics‑rich foundation for capturing machining know‑how as typed entities (features, tools, operations, machines) and relations (e.g., requiresTool, precedes, causes) \cite{xiao2023knowledge,wen2023systematic,hoang5097516vector}. For CAPP, Wang et al. generated a feature‑machining KG and proposed a cosine‑similarity retrieval that chose industry‑valid hole‑machining schemes with 0.8450 similarity to expert templates~\cite{wang2025machining}. Guo et al. organized historical process routes by feature topology and machine capability, trimming part‑routing time from 50–80 min to approximately 15 min~\cite{guo2023knowledge}. Beyond planning, KGs have achieved >90\% accuracy in root‑cause diagnostics of rotary machinery even under missing sensor data by reasoning over causal chains~\cite{cai2024research}. Despite these gains, KG deployment still depends on laborious, manual curation: (1) domain experts must annotate source texts, validate triples, and periodically inject new machining knowledge; (2) Simple subject–predicate–object triples also struggle to encode context such as tolerance stack‑ups, fixture constraints, or process‑chain rationale for nuance components. For example, a given connection between the tool used to create thin walled geometry might omit the dependency between the optimal tool geometry corresponding the the feature size and specific tolerances required for that specific part.

Figure~\ref{fig:why} showcases these limitations by contrasting (1) a stand‑alone knowledge graph that offers traceable but context‑starved reasoning, returning no answer when crucial relations such as material‑specific clearance or over‑size rules are absent, and (2) a stand‑alone LLM that confidently hallucinates an incorrect 8 mm recommendation, revealing the need for a hybrid KG and LLM architecture that is both provenance‑aware and context‑adaptive. This research introduces Augmented Retrieval Knowledge Network Enhanced Search and Synthesis (ARKNESS), a hybrid framework that combines LLMs with semantically enriched, domain-specific KGs that provides answers that are grounded in validated information ensuring that recommendations for machining process planning queries are both reliable and contextual relevant. In addition, by grounding LLM prompts in these rich subgraphs, ARKNESS can deploy a compact parameter model on-premise with accuracy on par with, or exceeding, much larger cloud LLMs, slashing compute and deployment overhead. The main benefits of this paper are:

\begin{enumerate}
    \item \textbf{A model‑agnostic, KG‑augmented CAPP assistant.} We present a retrieval‑augmented generation pipeline that couples any large language model with a multi‑relational machining knowledge graph. A graph‑aware prompt compiler injects provenance‑linked triples into the context window, yielding numerically precise and explainable process‑planning answers.
    \item \textbf{Self‑supervised KG distillation from technical corpora.} A zero‑shot, GPT‑based entity–relation extractor converts heterogeneous machining documents (e.g., PDF manuals, specification sheets, and NC code comments) into contextualized triples, eliminating the manual curation bottleneck that has constrained previous CAPP‑oriented KGs.
    \item \textbf{Lightweight, on‑prem hallucination suppression.} On a 155‑query benchmark, ARKNESS paired with a smaller LLM model matches or surpasses much larger cloud LLMs while reducing hallucinations, demonstrating a privacy‑preserving path to real‑time shop‑floor deployment.
\end{enumerate}

The rest of the paper is as follows: Section~\ref{bkgd} provides relevant knowledge graph and large language modeling research in manufacturing planning contexts, Section~\ref{methodology} provides the implementation of ARKNESS for automatic graph creation for retrieval and answering, Section~\ref{experiment} describes the experimental setup, Section~\ref{results} presents the experimental results, and Section~\ref{conclusion} concludes the paper.

\section{Research Background}\label{bkgd}
\subsection{Large Language Models in Machining}

LLMs are rapidly transitioning from research curiosities to core enablers of AI‑driven production, thanks to their capacity to parse free‑form instructions, fuse heterogeneous context, and return actionable, domain‑specific guidance. A recent survey by Li et al. categorized their early penetration into the manufacturing stack—spanning generative CAD modeling, bio‑process recipe design, robot path planning, and vision‑based quality control~\cite{li2024large}. Within subtractive manufacturing, the most direct application is natural‑language‑to‑G‑code translation. Šket et al. evaluated commercial ChatGPT models for 3-axis G-code generation testing GPT-3.5 and GPT-4 in three phases including independent G-code generation, interpretation of the generated G-code, and detecting and simplifying errors~\cite{vsket2024enhancing}. Their results showed promise of implementing LLMs for G-code generation with GPT-4 producing more correct toolpaths but is severely limited to simple operations such as drilling. Additionally, their method relied heavily on user input to align the LLMs for generation which can lead to increased downtime if deployed in manufacturing environments. Jeon et al. expanded upon just G-code generation by developing ChatCNC that integrated various LLM agents with real-time CNC machining data~\cite{jeon2025chatcnc}. This allowed users receive context-aware answers regarding the status of their 3-axis CNC machines such as spindle load at specific instances or times; their method achieved an accuracy of 93.3\% in queries requiring complex data inference such as production tracking showcasing applications in analyzing data recorded in the manufacturing pipeline. Despite the high accuracy, the authors acknowledged that the model resulted in failures when encountered with missing context or the inability to retrieve information from their database. This suggests a more in depth search beyond conventional retrieval methods such as defined SQL database structures is required to supplement existing knowledge gaps.

Beyond prompting and traditional retrieving techniques, researchers have also introduced domain-specific fine-tuning to better align a LLMs responses with domain-specific knowledge. Rosati et al. finetuned Llama 3 for industrial applications achieving uplifts in average ROUGE-1 F1-score from 0.164 to 0.314 when trained on user manuals of a 360-degree camera~\cite{rosati2024improving}. Wang et al. finetuned GPT-3.5 for air craft maintenance outperforming general GPT-3.5 and its upgraded counterpart GPT-4.0~\cite{wang2024ontology}. In areas specific to machining, Soundararajan et al. developed CNCGPT aimed for on-site CNC operator assistance by finetuning GPT-3.5 Turbo on machine-specific data, operational instructions, and troubleshooting assertions~\cite{kanimozhi2024explorative}. Results showed an increase in factual correctness when using ROGUE-L scores from 0.296 to 0.692 before and after finetuning, respectively. Despite the validity of their method, there are some challenges that might prevent further deployment on factory floors. One such issue is the amount of resources and training time required as the authors only trained around thirty query-response examples. In real world scenarios, there might be hundreds to thousands of scenarios and responses which can severely limit scalability during training. Moreover, as new scenarios continually emerge, the system must be frequently retrained to stay current which can further strain computational resources and complicates maintenance. There is also the challenge of security regarding production processes and information contained within. The authors used commercial LLMs which operate on external, cloud-based platforms where data flows and storage may raise concerns about data privacy and intellectual property protection. This reliance on third-party systems increases the risk of unauthorized access or data leakage, making it imperative to implement solutions that explore on-premise solutions to ensure that proprietary manufacturing data remains strictly confidential.

\subsection{Knowledge Graphs in Machining}
The ability to reuse and implement machining knowledge in a structured way provides not only enhanced consistency across planning and decision making but also ensures existing knowledge gaps are mitigated. Xiao et al. reviewed how computer-aided process planning involving knowledge graphs can benefit from reduced labor costs, shortened production cycles, and more intelligent use of existing information~\cite{xiao2023knowledge}. The authors analyzed key steps of implementing knowledge graphs from process knowledge representation to process knowledge graph construction and validation, showcasing how these methods can overcome traditional CAPP by reducing excessive manual interventions, and increasing flexibility and generalization. This implementation of process knowledge was shown in Wang et al. where they constructed a process knowledge graph for feature-based machining to automate machining scheme selection~\cite{wang2025machining}. Through use of an improved cosine-similarity formula for machining scheme selection, they achieved a similarity score of 0.8450 closely matching existing mature schemes implemented in industry for a typical shell part composed of 6 holes. By using their method, Wang et al. argued that the recommended machining steps would reduce tool load and increase both part quality and machining safety. In a similar study, Guo et al. created a knowledge graph for process route reuse by organizing historical process plans using part feature topology and machine capabilities~\cite{guo2023knowledge}. When introducing a new part, the authors determine the process route that best aligns with its feature topology through a similarity check with existing process routes. Their case study on a shaft part resulted in completing the overall machining process route in about 15 minutes, down from the typical 50 to 80 minutes, significantly enhancing efficiency.

Another major application of machining knowledge graphs involves fault diagnosis and maintenance of CNC equipment. Manufacturing systems generate heterogeneous data from sources such as sensors, logs, and maintenance reports that if left in isolation or silos can be difficult for diagnosis if problems arise. Knowledge graphs offer a unified representation by connecting physical components, signals, and failure modes in a network of cause-effect and part-whole relationships. Qiu et al. tackled this "data island" problem by constructing a multi-layered knowledge graph that integrated sensor data and domain knowledge, enabling the automatic identification of health changes in the X-axis ball screw drive system on a vertical milling tool (model XHK-5140) over 201 days, while similarity-based reasoning over the graph quantified the deviation from a healthy baseline~\cite{qiu2022novel}. Building on this idea, Cai et al. introduced a multi-level fault diagnosis KG for rotating machinery combining hierarchical knowledge of subsystem states with Bayesian reasoning, offering probabilistic inference across the graph to pinpoint root causes~\cite{cai2024research}. Notably, even with missing sensor outputs, their method achieved 91.1\% diagnostic accuracy by leveraging the relationships between related symptoms and outperformed traditional rule-based fault diagnosis. 

Despite the considerable progress in implementing knowledge graphs, there remain challenges in deploying them at scale for machining and industry. These aforementioned methods continue to depend heavily on extensive manual curation. The process of extracting, validating, and continuously updating the domain knowledge requires significant human knowledge and intervention, making it difficult to scale rapidly or adapt quickly to new information. This labor-intensive approach not only increases the risk of inconsistencies or omissions but also impedes real-time responsiveness in dynamic manufacturing environments. Furthermore, the conventional reliance on simple triple-based representations constrains the ability to capture the full contextual richness necessary for nuanced decision-making in machining processes. Such representations often fall short in encoding the detailed circumstances, rationale, and intricate relationships that underlie complex manufacturing operations. As a result, they may fail to fully express how various machining operations interrelate or why specific process choices are chosen. This would thus lead to a superficial understanding of the underlying mechanics and process dependencies. These limitations highlight the need for more flexible and richly contextualized approaches in knowledge representation to support the evolving demands of modern machining applications.

\section{Research Methodology}\label{methodology}
\begin{figure}[h!]
    \centering
    \includegraphics[width=0.8\textwidth]{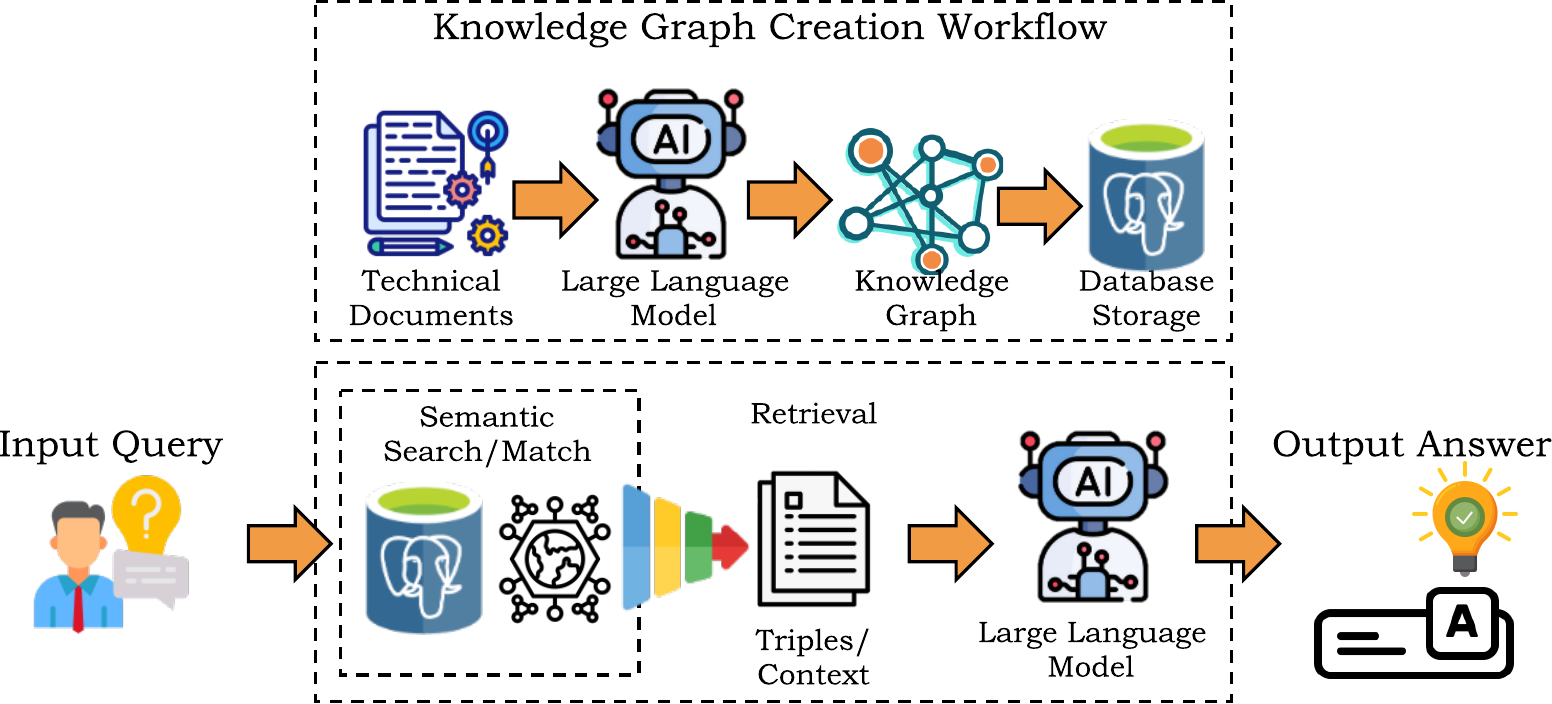}
    \caption{Overview of the framework with incorporating knowledge graph with large language model, illustrating the process from graph creation and storage to semantic search, match, and retrieval, and finally to language model response.}
    \label{fig:overview}
\end{figure}

This section details the two main components of the ARKNESS framework with 1) Knowledge graph construction and 2) Graph transversal, retrieval and LLM response. Figure \ref{fig:overview} provides the general overview of the two components.

\subsection{Knowledge Graph}
A knowledge graph can be considered as a set of tuples $\mathcal{G} = \{(v,r,w)\}$ where $v$ and $w$ are vertices (entities) from a set of vertices $\mathcal{V}$ and $r$ is an edge (relationship) from a set of relationships $\mathcal{R}$. The set of tuples is commonly known as triples denoted as (\textit{head}, \textit{relation}, \textit{tail}) representing $v$, $r$, and $w$, respectively.

\subsection{Automated Graph Construction}\label{sec:graph_construct}
The knowledge graph construction for ARKNESS is mainly done using textual data from documents relevant to the users chosen domain. Due to the heterogeneous nature of document types and file formats (e.g., .docx, .pdf, .pptx), preprocessing was first done to extract the textual information from these different sources. This was achieved through the Docling~\cite{auer2024docling} Python package which efficiently parses through the documents and extracts the raw text subsequently exporting the content as Markdown files for further processing.

Document splitting was done through parsing out each individual paragraph. Given a document with $p$ individual paragraphs, each paragraph was given to a LLM to extract entities and relations forming $T$ triples. Specific instructions were given to guide the LLM to output structured information given below:
\begin{lstlisting}
-Goal-
Extract structured triples directly from the input text in the following format:

ENTITY_1, RELATIONSHIP_TYPE, ENTITY_2, "RELATIONSHIP_DESCRIPTION"

-Steps-
1. Read the input text carefully to identify:
   - Key entities: Concepts, systems, technologies, or processes central to the text.
   - Relationships: Clear actions or connections described in the text that link two entities.
   - Descriptions: Verbatim or paraphrased descriptions from the text that explain the relationship.

2. For each relationship, construct a triple:
   - ENTITY_1: The primary concept or entity initiating the relationship.
   - RELATIONSHIP_TYPE: The action or connection type as described in the text.
   - ENTITY_2: The target concept or entity affected by ENTITY_1.
   - RELATIONSHIP_DESCRIPTION: A concise description of the relationship directly sourced from the input text.

3. Each triple must be clear and in this format:
   ENTITY_1, RELATIONSHIP_TYPE, ENTITY_2, "RELATIONSHIP_DESCRIPTION"

4. Use the original text verbatim where possible for the description, ensuring accuracy. Avoid adding external interpretations or explanations.
\end{lstlisting}


This process was repeated for each document chosen creating $G_D$ subgraphs. The purpose of including the 'Relationship Description' in the triplet information is to retain the relevant contextual information by which the triplet was created. This removes ambiguity and often missing information found in traditional knowledge graphs. All subgraphs were then stored into respective text files which were then combined together for further processing.

A hierarchical database was then created using PostgreSQL to efficiently store and retrieve entities, relations, and their associated context. Specifically, Python and the Psycopg library was used to interface with the database where during initialization three primary tables are created: (1) subjects table: stores unique entity names with a serial primary key, (2) relations table: associates each subject with its corresponding relationships where each entry references a subject's unique identifier, ensuring relationships are correctly linked, and (3) objects table: stores ending triplet entities along with corresponding contextual information. This structure can be also represented as follows. Let $S$ be the set of subjects (entities). For each subject $s \in S$, let $R(s)$ be the set of relations associated with $s$. For each relation $r \in R(s)$, let $O(r)$ be the set of ending objects linked to $r$, where each object is represented as a tuple $(v,c)$ with $v$ representing connected entities, and $c$ representing the corresponding context. The entire data base is thus defined by the set:
\begin{equation}
    \mathcal{D} = \{\langle s,r, (v,c)\rangle | s \in S, r \in R, (v,c) \in O(r)\}
\end{equation}
with a single triple entry represented as:
\begin{equation}
    \langle s,r, (v,c)\rangle
\end{equation}

\subsection{Graph Retrieval and Transversal}

\begin{figure}[h!]
    \centering
    \includegraphics[width=\textwidth]{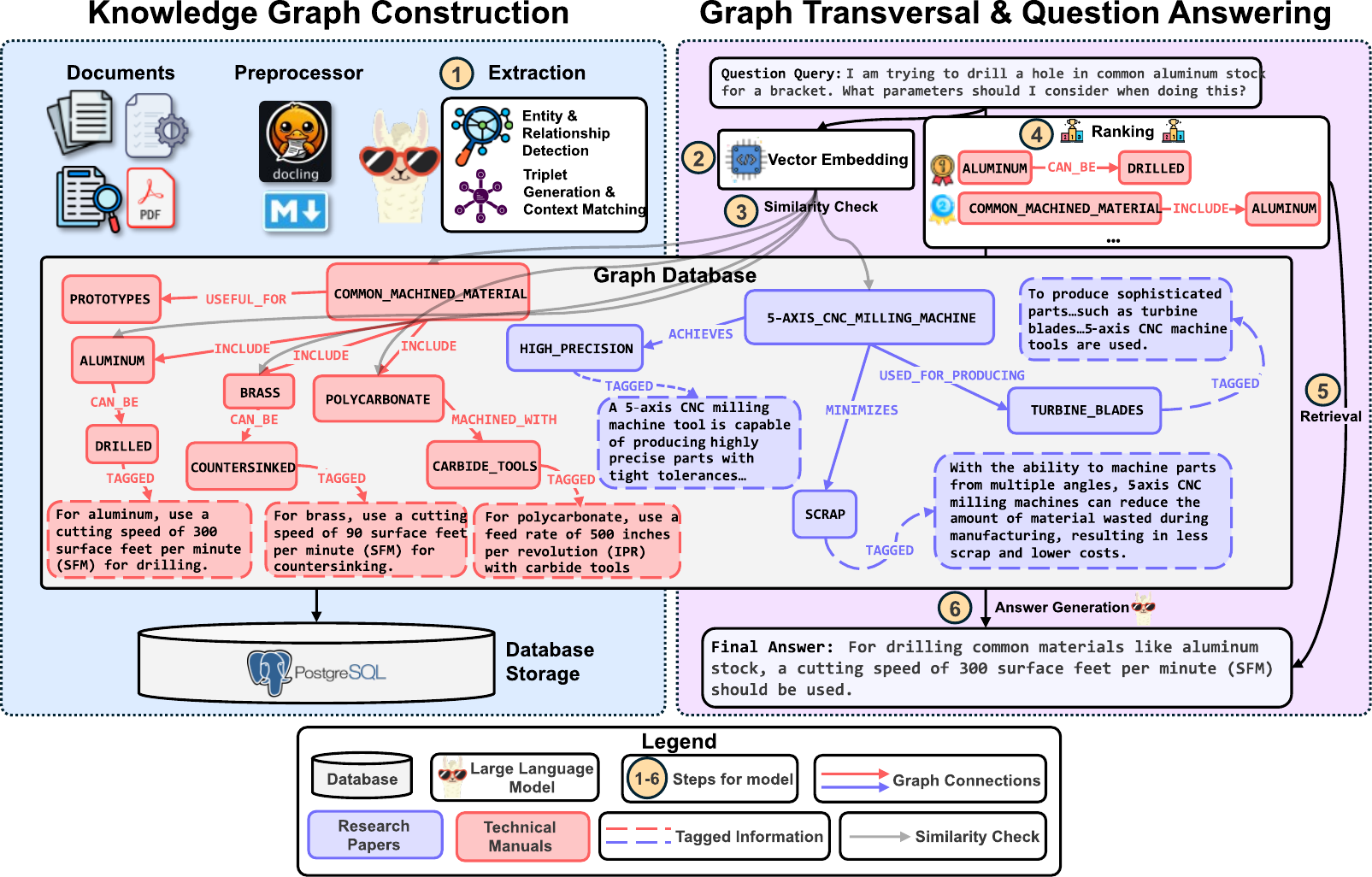}
    \caption{Overview of the knowledge graph construction and graph transversal for user queries. The left side depicts the automated graph construction process from document selection, preprocessing, entity-relationship extraction, triplet generation and context matching, and database storage. The right side depicts knowledge graph retrieval given a user query by embedding the query to get a vector representation, performing a similarity check over the triples in the graph, and ranking the most relevant triples to give to the large language model for answer generation.}
    \label{fig:indepth}
\end{figure}
Figure \ref{fig:indepth} provides an indepth illustration of how triplets are compared against a query and retrieved to be given to a LLM. Once the database is created and stored, the information contained within is incorporated to provide additional context and information to anchor subsequent questions. This is achieved through retrieving the relevant triplets and context from the knowledge graph. The retrieval process begins by first encoding a given natural language query, or question, into a vector $q \in \mathcal{R}^d$. This is achieved using a chosen semantic embedding model $f: \mathrm{Text} \rightarrow \mathcal{R}^d$. Similarly, each triplet $t_i$ from the database is encoded into embedding $e_i$ using the same semantic embedding model where $f: t_i \rightarrow \mathcal{R}^d$. The similarity between the query and each triple is then calculated using cosine-similarity according to:
\begin{equation}
    \mathrm{sim}(q,e_i) = \frac{q \cdot e_i}{\|q\| \|e_i\|}
\end{equation}
This similarity score provides an evaluate of how semantically similar the query given is to each of triplet in the knowledge graph allow one to ascertain its relevance.

Given the potentially massive scale of knowledge graphs, which can reach hundred of thousands or more connections, a candidate pool $C$ is constructed by selecting indices corresponding to a chosen a top-$K$ triples with the highest similarity scores $ s_i = \mathrm{sim}(q,e_i)$. Having only the highest similarities chosen allows the most semantically relevant triples to be selected for further processing. The candidate pool is thus defined as:
\begin{equation}
    C = \{ i \in \{1,2,...,K\}\}~\text{with}~s_{i_1} \geq s_{i_2} \geq \dots \geq s_{i_k}
\end{equation}

As mentioned previously, because a knowledge graph can contain hundred of thousands or even millions of connections, relying solely on the top-$K$ triples could overlook additional relevant triples connected to them. To explore the extended neighborhood of these triples, we transverse through the graph by performing a beam search expansion. Let $E(i)$ denote the set of adjacent embedded triples indexed by $i$ (i.e., those sharing a node with triple $i$). We conduct a beam search recursively over a chosen maximum expansion depth $d_{max}$. Here higher depths allows one to explore deeper through the graph allowing more information to be retrieved. For an initial depth $d = 0$, we first set the initial candidates as $C_0 = C$. For each subsequent depth $d \geq 1$ we find additional candidates using:
\begin{equation}
    C_d = \bigcup_{i \in C_{d-1}} \text{Top-}b\Bigl(\{ j \in E(i) \mid \mathrm{sim}(q,e_j)\}\Bigr)
    \label{depth}
\end{equation}
where Top-$b(\cdot)$ selects the chosen number of $b$ candidates, also known as beam width, with the higher similarity scores relative to the embedded query $q$. Here, beam width is a chosen parameter where higher values of $b$ allows the model to explore a wider set of candidates to expand from for each step. This can potentially capture more semantically relevant triples missed in the initial top-$K$. Conversely, lower values of $b$ restricts the search to the more promising candidates which reduces computational overhead at the expense of possibly missing relevant connections that lie outside the top $b$ candidates. In order to avoid re-processing triples, a set of visited triples $E_v$ is maintained such that the recursive expansion only considers new nodes defined as:
\begin{equation}
    C_{d+1} = \bigcup_{i \in C_d \setminus V} \text{Top-}b\Bigl(\{ j \in E(i) \}\Bigr)
\end{equation}
After recursively expanding through the chosen depth and beam size, the final set of triples retrieved is defined by:
\begin{equation}
    C^* = \bigcup_{d=0}^{d_{\max}} C_d
\end{equation}
As a final step after the final set of candidate triples are chosen, the corresponding context $c$ from the database is retrieved as well. This creates a set $\mathcal{C}$ of relevant information:
\begin{equation}
    \mathcal{C} = \{ \langle s_i,r_i,(o_i,c_i) \rangle : i \in I\}
\end{equation}

\subsection{Large Language Model Generation}
The last step of answering the query $q$ involves prompt construction where the information retrieved from the knowledge graph is given to the large language model to help its answering. This is done by defining a set of system instructions that designate that the LLM is design to answer questions incorporate both its own knowledge and the given retrieved knowledge. For example, we define the system instructions as:
\begin{equation}
    I_{\text{sys}} = \text{\parbox{0.8\linewidth}{'You are designed to help answer questions using retrieved knowledge. Not all knowledge given need to be used but focus on the most important information. Remember this knowledge, if there is any, to help your decision making:'}}
\end{equation}
Then the final prompt $P$ given to the LLM is constructed by concatenating the system instructions, query, and retrieved knowledge graph information:
\begin{equation}
    P(q,\mathcal{C}) = I_{\text{sys}} \oplus \mathcal{C}\oplus q
\end{equation}
The final answer $A_{\text{LLM}}$ is then generated by the LLM:
\begin{equation}
    A_{\text{LLM}} = \text{LLM}\bigl(P(q,\mathcal{C}))\bigr)
\end{equation}



\section{Experimental Design}\label{experiment}

\subsection{Knowledge Graph}
To test the efficacy of ARKNESS, we first created our machining knowledge graph composed of research and technical documents spanning 5-axis milling capabilities, industrial robotic integration in CNC machining, sustainability in CNC machining, CNC programming and process planning, and fundamentals of CNC machining. By exposing ARKNESS to a spectrum of complexity levels, domain‐specific terminologies, and the assembled documents enables a comprehensive evaluation of the model’s CNC machining knowledge. Each document was processed using the automated graph‐construction pipeline detailed in Section \ref{sec:graph_construct} using GPT-4o~\cite{hurst2024gpt}. An example regarding the output given the text information is given in the following box~\cite{soori2024review}. Descriptions of each documents chosen are listed in Table \ref{tab:documents} along with the total number of entities, relations, and triples after processing. In total, the full knowledge graph contains 4329 triples, 6659 unique entities, and 1251 unique relations providing a relatively large graph to test to test ARKNESS' capacity robust inference, and accurate knowledge retrieval across diverse CNC‑machining scenarios. An example of the graph-construction output given text input is given below:
\begin{lstlisting}
Input:
A 5-axis CNC milling machine tool is a sophisticated piece of equipment used for precision machining of complex parts with high accuracy and consistency. It has five axes of motion that can move the cutting tool in various directions to create intricate shapes and contours. The machine is designed to perform a wide range of operations, including milling, drilling, cutting, and engraving, using multiple axes of movement. A 5-axis CNC milling machine tool can create highly complex geometries using five axes of motion during machining operations that which can be difficult or impossible to be implemented by using traditional 3-axis machine.

Output:
5-AXIS CNC MILLING MACHINE TOOL, is used for, PRECISION MACHINING, "A 5-axis CNC milling machine tool is a sophisticated piece of equipment used for precision machining of complex parts with high accuracy and consistency."
5-AXIS CNC MILLING MACHINE TOOL, has, FIVE AXES OF MOTION, "It has five axes of motion that can move the cutting tool in various directions to create intricate shapes and contours."
5-AXIS CNC MILLING MACHINE TOOL, performs, "MILLING, DRILLING, CUTTING, AND ENGRAVING", "The machine is designed to perform a wide range of operations, including milling, drilling, cutting, and engraving, using multiple axes of movement."
5-AXIS CNC MILLING MACHINE TOOL, creates, HIGHLY COMPLEX GEOMETRIES, "A 5-axis CNC milling machine tool can create highly complex geometries using five axes of motion during machining operations that which can be difficult or impossible to be implemented by using traditional 3-axis machine."
TRADITIONAL 3-AXIS MACHINE, limits, GEOMETRY COMPLEXITY, "Machining operations that can be difficult or impossible to be implemented by using traditional 3-axis machine."
\end{lstlisting}

\begin{table}[ht!]
    \centering
    \caption{Overview of source documents and their corresponding knowledge graph sizes.}
    \scalebox{.95}{%
    \begin{tabular}{%
        p{4cm}                                     
        p{8cm}                                     
        >{\raggedright\arraybackslash}p{4cm}       
    }
        \toprule
        \textbf{Document} 
        & \textbf{Description} 
        & \textbf{Knowledge Graph Size} \\
        \midrule
        A Review in Capabilities and Challenges of 5‑Axis CNC Milling Machine Tool Operations~\cite{soori2024review}
        & This review surveys recent advances and challenges in 5‑axis CNC milling, including error modeling and compensation, toolpath and process optimization, virtual machining systems, tool wear and temperature prediction, and sustainability considerations.
        & \makecell[tl]{%
            \# of triples: 464 \\ 
            \# of unique entities: 654 \\ 
            \# of unique relations: 199%
          } \\
        \midrule
        Design and development of a CNC machining process knowledge base using cloud technology~\cite{ye2018design}
        & This paper presents a cloud‑based CNC machining process knowledge base that maps STEP‑NC to an OWL ontology and leverages Hadoop’s HBase for scalable storage, MapReduce driven querying, and SWRL‑based reasoning to enable intelligent, high‑throughput process planning.
        & \makecell[tl]{%
            \# of triples: 559 \\ 
            \# of unique entities: 828 \\ 
            \# of unique relations: 291%
          } \\
        \midrule
        Exploring the Application of Industrial Robots in CNC Machining~\cite{guo2023exploring}
        & This paper explores the deployment of industrial robots in CNC machining. It details robot selection, workflow and control program design, joint and transmission mechanisms, and drive system choices to automate loading/unloading and enhance operational flexibility, precision, and efficiency.
        & \makecell[tl]{%
            \# of triples: 98 \\ 
            \# of unique entities: 151 \\ 
            \# of unique relations: 57%
          } \\
        \midrule
        Fundamentals of CNC Machining~\cite{autodesk2014cnc}
        & This guide provides a practical introduction to CNC machining fundamentals covering shop safety, tooling, coordinate systems, programming and operation of mills and lathes, 2D/3D toolpaths, workholding examples, and best practices for prototype and short‑run production.
        & \makecell[tl]{%
            \# of triples: 1902 \\ 
            \# of unique entities: 2966 \\ 
            \# of unique relations: 512%
          } \\
        \midrule
        Innovative Approaches to Sustainable CNC Machining: A Machine Learning Perspective on Energy Optimization~\cite{nugrahanto2023innovative}
        & This paper develops a machine‑learning framework for sustainable 5‑axis CNC milling by combining per‑axis power monitoring with Taguchi experimental design and tree‑based regression to model and predict energy consumption.
        & \makecell[tl]{%
            \# of triples: 278 \\ 
            \# of unique entities: 452 \\ 
            \# of unique relations: 183%
          } \\
        \midrule
        Integration of Taguchi and PROMETHEE for CNC Milling Machining Parameter Optimization~\cite{ihsan2024integration} 
        & This paper integrates Taguchi’s orthogonal‑array experimental design with the PROMETHEE multi‑criteria ranking method to optimize spindle speed, feed rate, and depth of cut for CNC milling of AA6061.
        & \makecell[tl]{%
            \# of triples: 227 \\ 
            \# of unique entities: 366 \\ 
            \# of unique relations: 74%
          } \\
        \midrule
        Research on CNC programming and machining process based on CAD/CAM technology~\cite{zhang2024cnc}
        & This paper introduces a CAD/CAM‑based CNC programming framework that uses Bayesian process‑skeleton mapping to link part features with reusable macro‑processes and applies residual‑height trajectory generation and nonlinear‑error compensation to optimize multiaxis toolpaths.
        & \makecell[tl]{%
            \# of triples: 269 \\ 
            \# of unique entities: 443 \\ 
            \# of unique relations: 167%
          } \\
        \midrule
        Review on Design Research in CNC Machine Tools Based on Energy Consumption~\cite{wu2024review}
        & This review surveys global advances in modeling, designing, and evaluating CNC machine tools from an energy‑consumption perspective.
        & \makecell[tl]{%
            \# of triples: 270 \\ 
            \# of unique entities: 470 \\ 
            \# of unique relations: 140%
          } \\
        \midrule
        Robotical Automation in CNC Machine Tools, A Review~\cite{soori2024robotical}
        & This review surveys robotics‑driven advances in CNC machining. Automated material handling and tool changing to adaptive control, quality inspection, data analytics, and collaborative robots are evaluated for their impacts on efficiency, precision, and safety.
        & \makecell[tl]{%
            \# of triples: 262 \\ 
            \# of unique entities: 380 \\ 
            \# of unique relations: 129%
          } \\
        \bottomrule
    \end{tabular}%
    }
    \label{tab:documents}
\end{table}

\subsection{Models}
To rigorously assess how model choice shapes our framework’s performance, we evaluate a diverse set of both open‑source and closed‑source large language models across multiple parameter scales. For open sources models, Llama 3.2 3B Instruct~\cite{meta2024llama32}, LLama 3.1 8B Instruct~\cite{grattafiori2024llama}, and Qwen 2.5 7B Instruct~\cite{yang2024qwen2} were chosen due to their ease of access and relatively low computational requirements; here B represents the number of parameters in billions. By spanning 3B, 7B, and 8B parameter tiers, we can isolate how model capacity and design choices interact with our knowledge graph augmentation.  Evaluating these models across a broad spectrum of parameter scales enables us to quantify how the integration of supplementary knowledge graph information influences overall performance. To establish an upper bound on attainable performance, we also benchmark against state‑of‑the‑art closed sourced models including GPT-4o~\cite{hurst2024gpt}, its smaller GPT-4o-mini, Gemini 2.0 Flash~\cite{deepmind2025geminiflash2}, and Gemini 2.0 Flash-Lite~\cite{deepmind2025geminiflashlite}. These comprehensive models enables us to quantify precisely how supplemental structured knowledge closes the gap between open source baselines and leading proprietary offerings.

\subsection{Question Category}
Following knowledge graph construction, we devised two question formats namely, multiple choice and open ended to evaluate each model’s capabilities. Multiple choice questions present a controlled benchmark for retrieval precision, enabling objective measurement and direct comparison across models, while open‑ended questions simulate real‑world case studies by challenging models to perform generative synthesis and coherently integrate the retrieved information into comprehensive answers. For each question format, two categories of questions were created namely: content specific and machining specific. Content specific questions refer to the information written in the documents that do not require quantitative analysis. Machining specific questions refer to questions that require quantitative precision and decision‑making based on numerical parameters. An example of each question in multiple choice format is given below:

\begin{tcolorbox}[myquestionbox,title=Content Specific Question]
Which statement best describes a key distinction between 3‑axis and 5‑axis CNC milling machines?\\[4pt]
A. 3‑axis machines are used for metals, while 5‑axis machines are used exclusively for plastics.\\
B. 5‑axis machines include tilt and rotation of the workpiece or tool, in addition to X, Y, and Z motion.\\
C. 3‑axis machines are larger and require more floorspace than 5‑axis machines.\\
D. 5‑axis machines do not allow any vertical movement while 3‑axis machines do.[4pt]
\end{tcolorbox}

\begin{tcolorbox}[myquestionbox,title=Machining Specific Question]
For drilling operations on stainless steel (303), what feed (in inches per revolution) is recommended?\\[4pt]
A. 0.0005 in/rev\\
B. 0.0010 in/rev\\
C. 0.0020 in/rev\\
D. 0.0030 in/rev
\end{tcolorbox}
\noindent GPT-4o was implemented to generate each question from the documents chosen with a total of 65 content specific multiple choice, 45 machining specific multiple choice, 104 content specific open ended, and 45 machining specific open ended.

\section{Experimental Results}\label{results}

\begin{figure}[h!]
    \centering
    \includegraphics[width=\textwidth]{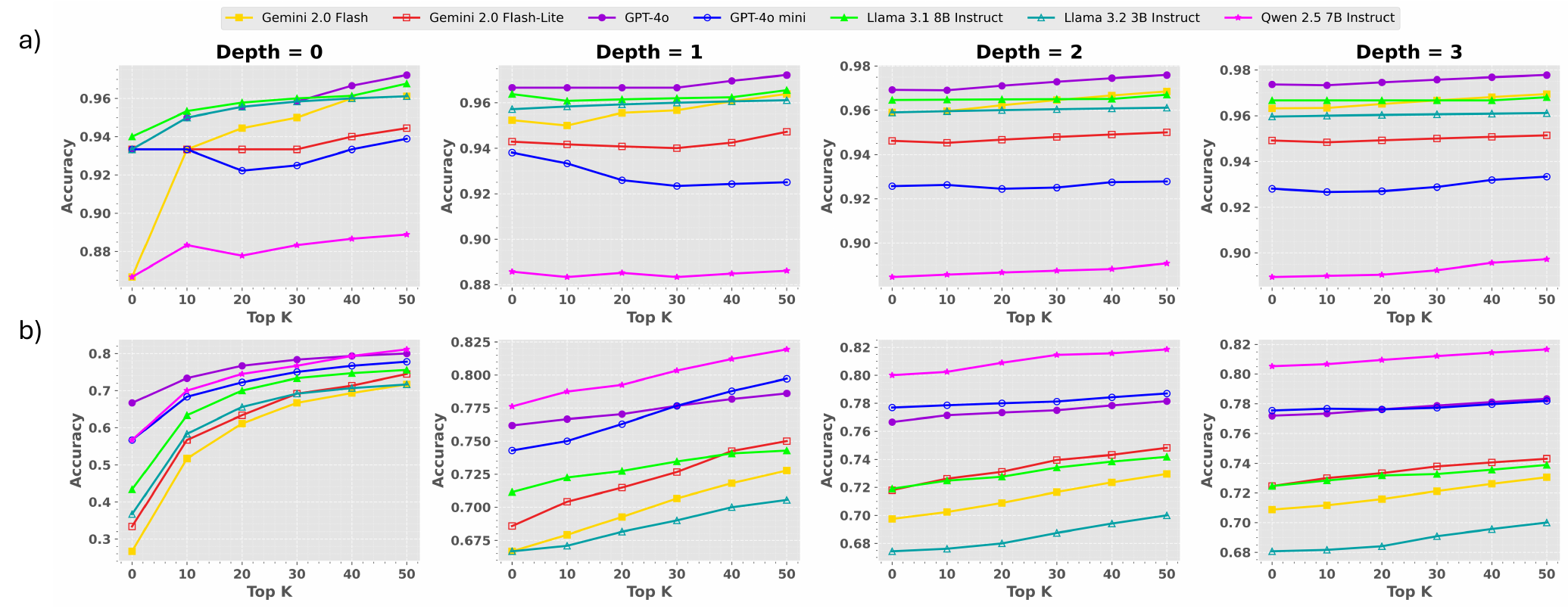}
    \caption{Graph of large language model model accuracy as a function of graph traversal depth and Top K triple retrieval for a) content specific queries and b) machining specific queries.}
    \label{fig:topk}
\end{figure}

ARKNESS is implemented using Python, PyTorch, and respective APIs for each of the chosen models. We initially explore how knowledge graph augmentation can enhance the performance of large language models using controlled graph traversals as defined by Eq. \ref{depth}. Figure \ref{fig:topk} shows experiments that vary the depth of these traversals and measure the effect on the accuracy of the LLM answers. In these experiments, we also focus on selecting the Top K highest scoring triples along with there corresponding context. Figure \ref{fig:topk}a presents results with multiple choice questions that were designated content specific. As shown, as we increase the number of Top K triples retrieved from the graph to give to the LLMs the general trend is that performance increases. This trend holds true across most depths tested as there is an exception where GPT‑4o mini at depth one experiences a decrease in performance, possibly because the additional graph-derived information is redundant or less informative, overwhelming the model's initial context. There is a notable increase for Gemini 2.0 Flash at depth 0 between no information given (Top K of 0) to Top K of 10 increasing from an accuracy of 0.867 to 0.933. It can also be observed that as the depth increases, the performance across the retrieved Top K stabilizes, indicating that for these types of questions a depth of 0 is most optimal since additional information does not yield further benefits. This stabilization implies that the initial retrieved context is already sufficiently rich, and incorporating extra graph-derived details beyond depth zero fails to enhance model accuracy and may even introduce superfluous noise. Compared to content-specific questions, we observe a significantly greater trend for quantitative machining specific questions, with performance improvements manifesting across all depths and Top K values as shown in Figure \ref{fig:topk}b). This indicates that additional graph-derived context plays a crucial role in accurately identifying the correct quantitative values for machine-specific queries, with stabilization around a depth of 2 demonstrating that these more challenging questions benefit from a deeper graph traversal to capture the necessary relevant information.
\begin{figure}[h!]
    \centering
    \includegraphics[width=.9\textwidth]{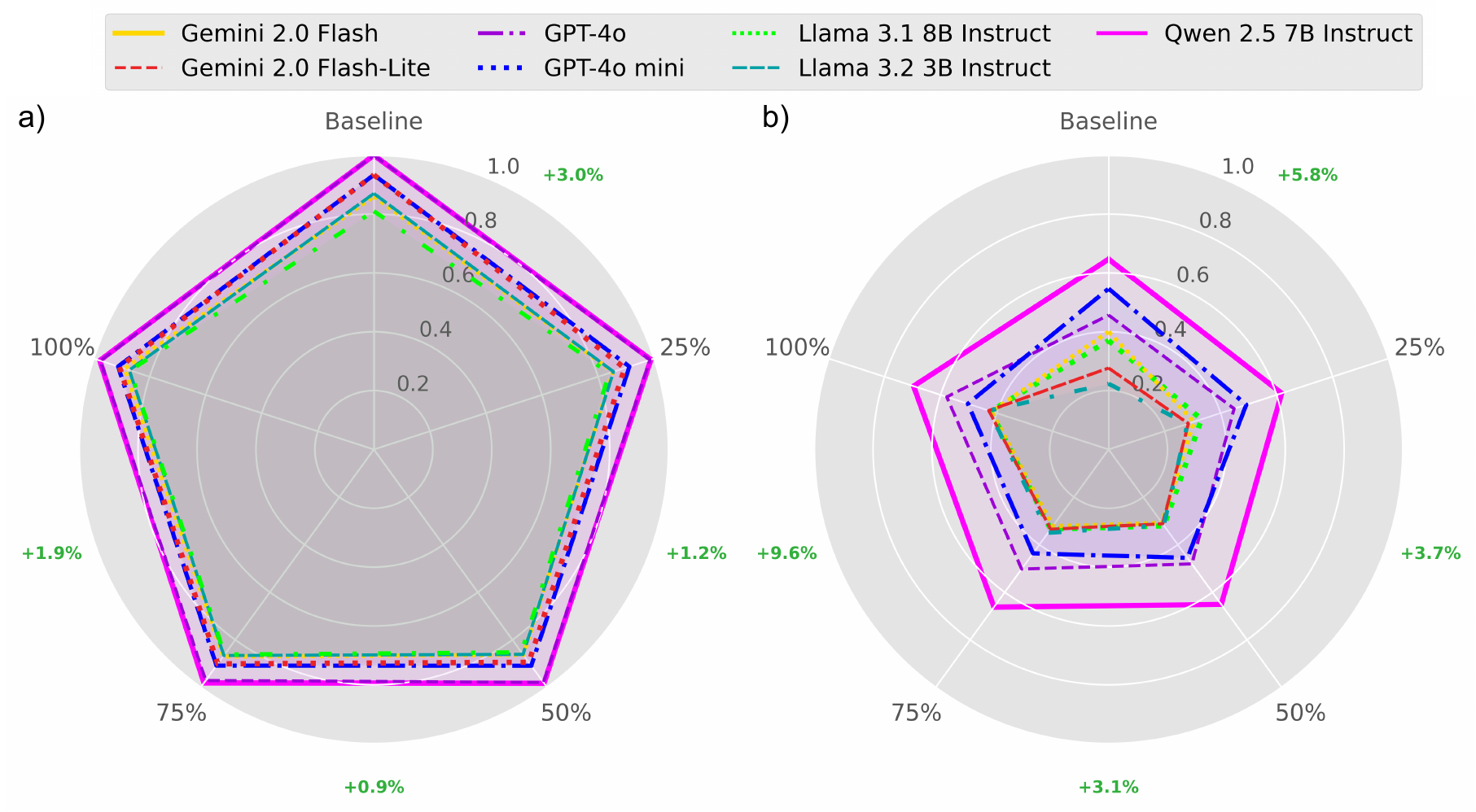}
    \caption{Radar plots of average large language model accuracy performance across knowledge graph completeness levels ranging from baseline (0\%) to full graph (100\%) for a) content Specific and b) machining specific multiple choice queries.}
    \label{fig:graphsize}
\end{figure}

Figure \ref{fig:graphsize} showcases our model under conditions of incomplete knowledge graphs to emulate real-world scenarios, where the available data may be partial or sparsely connected. This approach not only tests the robustness of the graph augmentation mechanism but also helps us understand how missing nodes and relations affect the retrieval of relevant information and, ultimately, the accuracy of the LLM responses. We again tested the same aforementioned LLMs under varying graph sizes, specifically, baseline with no graph, 25\%, 50\%, 75\%, and full graph of 100\%. We randomly dropped triples to achieve the chosen graph sizes and each model was tested 10 times for each respective graph size with the average reported on the plot and the green percent increase indicating the maximum increase across all models tested when increasing the graph size. Figure \ref{fig:graphsize}a presents the performance metrics for content-specific questions across various LLMs. The results indicate only a modest improvement when additional graph context is provided, suggesting that these questions can largely be addressed with the baseline contextual information available to each model. Notably, the maximum performance gain, an absolute increase of approximately 3\%, is observed when the graph size is augmented from the baseline to 25\%, demonstrating that while supplementary retrieved data offers some benefit, its impact remains relatively limited for content-specific queries. Figure \ref{fig:graphsize}b illustrates how expanding the graph size affects performance on the more specialized quantitative machining specific questions. Unlike the content specific queries in Figure \ref{fig:graphsize}a, these questions exhibit more pronounced gains as the graph grows beyond the baseline level indicated by the greater maximum increase across the models. When comparing closed source models with open sourced, GPT‑4o effectively leverages the additional graph context to improve performance by 12.1\% at a graph size of 100\% compared to baseline. Additionally, Gemini 2.0 Flash-Lite achieves a relatively high increase of 15.1 \% compared to baseline at a graph size of 100\%. In contrast, the open source Llama 3.2 3B Instruct shows an even greater relative gain of 20.3\% at graph size of 100\% compared to baseline, suggesting that these models, which may have a lower baseline performance, can be significantly boosted by enhanced external context. Notably, the significant improvements seen in the smaller open source models imply that in manufacturing environments, such as the factory floor, these models may be particularly effective when enhanced with additional contextual information. Their lower computational overhead and faster inference speeds make them especially suited for deployment in real-time industrial settings where cost efficiency and rapid responsiveness are critical.

\begin{table}[ht!]
\centering
\caption{Benchmark accuracy and F1‐score performance between baseline and with knowledge graph for each large language model. Absolute increases (or no change) are in \textcolor{green}{green}, and decreases are in \textcolor{red}{red}.}
\label{tab:benchmark}
\resizebox{0.8\textwidth}{!}{%
  \begin{tabular}{@{\hskip 5pt}l@{\hskip 5pt}c@{\hskip 5pt}c@{\hskip 5pt}c@{\hskip 5pt}c@{\hskip 5pt}c@{\hskip 5pt}c@{\hskip 5pt}}
    \toprule
    \multirow{2}{*}{Model} & \multirow{2}{*}{File} & \multicolumn{2}{c}{Baseline} & \multicolumn{2}{c}{Knowledge Graph} \\ 
    \cmidrule(lr){3-4} \cmidrule(lr){5-6}
      & & Accuracy & F1 Score & Accuracy & F1-Score \\ 
    \midrule
\multirow{2}{*}{Gemini 2.0 Flash}
    & Content Specific   & 0.867 & 0.535 & 0.933 \textcolor{green}{(+0.066)} & 0.569 \textcolor{green}{(+0.034)} \\
    & Machining Specific     & 0.267 & 0.152 & 0.517 \textcolor{green}{(+0.250)} & 0.224 \textcolor{green}{(+0.072)} \\ \midrule
\multirow{2}{*}{Gemini 2.0 Flash-Lite}
    & Content Specific   & 0.933 & 0.888 & 0.933 \textcolor{green}{(0.000)}   & 0.688 \textcolor{red}{(-0.200)} \\
    & Machining Specific     & 0.333 & 0.277 & 0.567 \textcolor{green}{(+0.234)} & 0.338 \textcolor{green}{(+0.061)} \\ \midrule
\multirow{2}{*}{GPT-4o} 
    & Content Specific   & 0.933 & 0.716 & 0.950 \textcolor{green}{(+0.017)} & 0.719 \textcolor{green}{(+0.003)} \\
    & Machining Specific     & 0.667 & 0.464 & 0.733 \textcolor{green}{(+0.066)} & 0.573 \textcolor{green}{(+0.109)} \\ \midrule
\multirow{2}{*}{GPT-4o mini}
    & Content Specific   & 0.933 & 0.906 & 0.933 \textcolor{green}{(0.000)}   & 0.901 \textcolor{red}{(-0.005)} \\
    & Machining Specific     & 0.567 & 0.391 & 0.683 \textcolor{green}{(+0.116)} & 0.549 \textcolor{green}{(+0.158)} \\ \midrule
\multirow{2}{*}{Llama 3.1 8B Instruct}
    & Content Specific   & 0.940 & 0.925 & 0.953 \textcolor{green}{(+0.013)} & 0.940 \textcolor{green}{(+0.015)} \\
    & Machining Specific     & 0.433 & 0.244 & 0.633 \textcolor{green}{(+0.200)} & 0.404 \textcolor{green}{(+0.160)} \\ \midrule
\multirow{2}{*}{Llama 3.2 3B Instruct}
    & Content Specific   & 0.933 & 0.718 & 0.950 \textcolor{green}{(+0.017)} & 0.712 \textcolor{red}{(-0.006)} \\
    & Machining Specific     & 0.367 & 0.319 & 0.583 \textcolor{green}{(+0.216)} & 0.543 \textcolor{green}{(+0.224)} \\ \midrule
\multirow{2}{*}{Qwen 2.5 7B Instruct}
    & Content Specific   & 0.867 & 0.580 & 0.883 \textcolor{green}{(+0.016)} & 0.614 \textcolor{green}{(+0.034)} \\
    & Machining Specific     & 0.567 & 0.371 & 0.700 \textcolor{green}{(+0.133)} & 0.463 \textcolor{green}{(+0.092)} \\
 \bottomrule
\end{tabular}
}
\end{table}

Table \ref{tab:benchmark} provides a benchmark comparison between baseline model performance and results with a minimal viable integration of knowledge graph information (i.e., a chosen Top K of 10 and depth of 0) for both multiple choice datasets. Results for both accuracy and F1-score are reported with the absolute increases, or no increase, between baseline and knowledge graph augmented outputs highlighted in green and any declines shown in red. As shown in the table, for content specific questions where baseline performance was already high, the improvements are very modest and range from 0.013 to 0.066 for accuracy. However, even this relatively small uplift indicates that the inclusion of the knowledge graph imparts valuable domain-specific insights that further refine the responses. Other the other hand, for the quantitative machining specific questions the results are far more substantial. For example, Gemini 2.0 Flash-Lite achieves an uplift of 0.250 in accuracy and 0.072 in F1-score. Similarly, magnitudes of improvement is shown with Llama 3.2 3B Instruct and Llama 3.1 8B Instruct having an uplift of 0.216 and 0.200 in accuracy, respectively. These improvements provide evidence that the structured, domain-specific context provided by the knowledge graph effectively addresses the deficiencies each LLM's internal representations. It should be noted that there is a notable decrease in F1-score of 0.200 for Gemini 2.0 Flash-Lite when testing on content specific questions. This decline may be attributed to the additional context provided by the knowledge graph, which, while generally beneficial, can increase the complexity of the input and lead to confusion in the model's internal representations and final decision making. In summary, despite some localized declines, likely due to increased input complexity, the integration of knowledge graph information consistently enhances performance across most models, providing significant improvements in domain-specific accuracy and reliability across both content specific and quantitative machining specific queries, and thereby demonstrating its critical value for augmenting LLM outputs in technical applications.

\begin{figure}[h!]
    \centering
    \includegraphics[width=\textwidth]{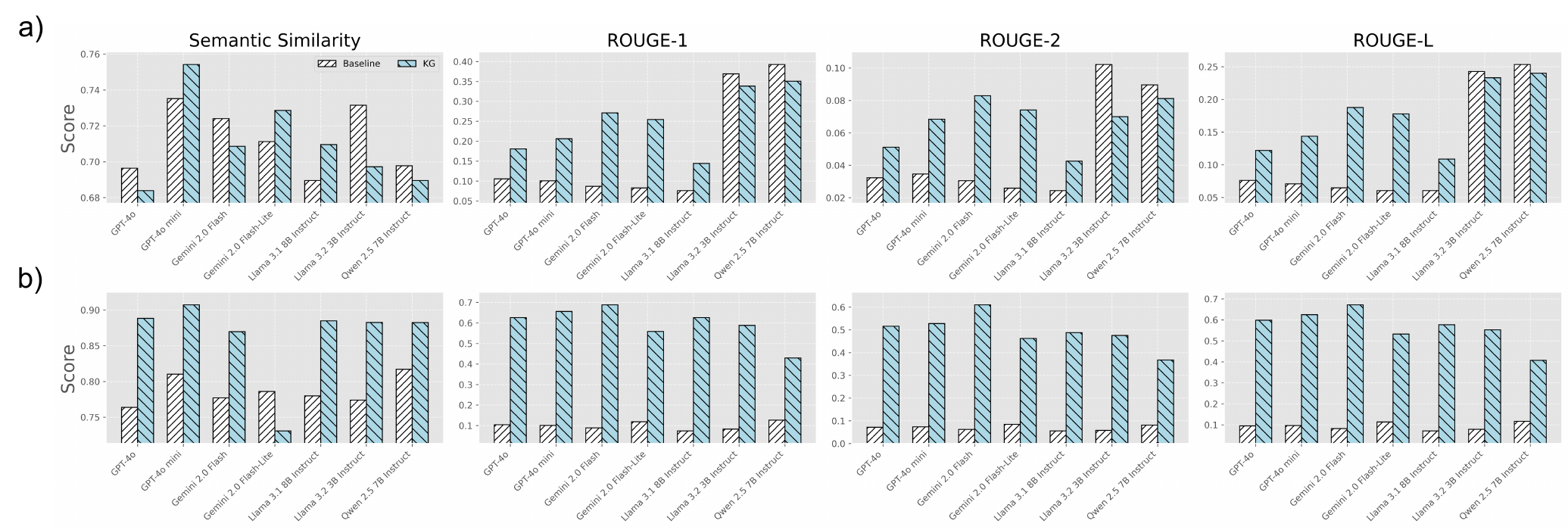}
    \caption{Average performance over 10 runs for evaluated large language models, comparing baseline outputs with those augmented by knowledge graph data. a) Shows performance for content specific questions, and b) depicts results for machining specific questions across semantic similarity, ROUGE-1, ROUGE-2, and ROUGE-L metrics.}
    \label{fig:semanticsim}
\end{figure}

\begin{table}[ht!]
\centering
\caption{Open ended responses from open source models Llama 3.2 3B Instruct, Llama 3.1 8B Instruct, and Qwen 2.5 7B Instruct regarding drill sizes for CNC machining. Grounded quantitative answers are highlighted in \textcolor{yellow}{yellow}, baseline values in \textcolor{red}{red}, and knowledge graph augmented values in \textcolor{green}{green}.}
\label{tab:drill_open}
\resizebox{\textwidth}{!}{
\begin{tabular}{p{3cm} p{5.33cm} p{5.33cm} p{5.33cm}}
\toprule
\textbf{Model / Question} & \textbf{What is the decimal equivalent of Drill Size 82?} & \textbf{What is the decimal equivalent of Drill Size 84?} & \textbf{What is the decimal equivalent of Drill Size 89?} \\
\midrule
\textbf{Grounded Answer} & 
The decimal equivalent of Drill Size 82 is \hlyellow{0.0125 inches}. 
& 
The decimal equivalent of Drill Size 84 is \hlyellow{0.0115 inches}. 
& 
The decimal equivalent of Drill Size 89 is \hlyellow{0.0091 inches}. \\
\midrule
\textbf{Llama 3.1 8B Instruct} & 
\textbf{Baseline:} We find that Drill Size 82 is actually equivalent to \hlred{0.082 inches}. This is because the number 82 represents the 82nd drill size in the series, and the decimal equivalent is calculated using a formula that takes into account the drill size and the series.
\newline
\newline
\textbf{Knowledge Graph:} The decimal equivalent of Drill Size 82 is \hlgreen{.0125 inches}.
& 
\textbf{Baseline:} The decimal equivalent of drill size 84 is \hlred{0.84 inches}, which can also be represented as a fraction, 84/64.
\newline
\newline
\textbf{Knowledge Graph:} The decimal equivalent of Drill Size 84 is \hlgreen{.0115 inches}.
& 
\textbf{Baseline:} The decimal equivalent of Drill Size 89 is approximately \hlred{0.139 inches}.
\newline
\newline
\textbf{Knowledge Graph:} The decimal equivalent of Drill Size 89 is \hlgreen{.0091 inches}. \\
\midrule
\textbf{Llama 3.2 3B Instruct} & 
\textbf{Baseline:} According to the standard drill size chart, Drill Size 82 corresponds to a diameter of \hlred{0.820 inches}. 
\newline
\newline
\textbf{Knowledge Graph:} To find the decimal equivalent of drill size 82, we need to look at the information provided. Drill size 82 corresponds to a decimal equivalent of \hlgreen{.0125 inches}.
& 
\textbf{Baseline:} Drill Size 84 refers to a drill bit with a diameter of 84\,mm. To convert this to a decimal equivalent, we simply need to express the diameter as a decimal number. In this case, the decimal equivalent of Drill Size 84 is \hlred{84.00 mm}.
\newline
\newline
\textbf{Knowledge Graph:} The decimal equivalent of Drill Size 84 is \hlgreen{.0115 inches}.
& 
\textbf{Baseline:} The decimal equivalent of Drill Size 89 is \hlred{0.89 inches, or 0.0089 inches}.
\newline
\newline
\textbf{Knowledge Graph:} The decimal equivalent of Drill Size 89 is \hlgreen{.0091 inches}. \\
\midrule
\textbf{Qwen 2.5 7B Instruct} & 
\textbf{Baseline:} ``Drill Size 82'' likely corresponds to a drill bit that measures 82 thousandths of an inch in diameter. This translates to \hlred{0.82 inches} when expressed as a decimal.
\newline
\newline
\textbf{Knowledge Graph:} The decimal equivalent of Drill Size 82 is \hlgreen{0.0125 inches}. 
& 
\textbf{Baseline:} The decimal equivalent of Drill Size 84 typically refers to a drill bit size in the United States customary system... Rounded to four decimal places, the decimal equivalent of Drill Size 84 is approximately \hlred{0.0119 inches}.
\newline
\newline
\textbf{Knowledge Graph:} The decimal equivalent of Drill Size 84 is \hlgreen{0.0115 inches}. 
& 
\textbf{Baseline:} The decimal equivalent of Drill Size 89, which typically refers to a drill bit size in the United States customary system, is \hlred{1/2 inch}.
\newline
\newline
\textbf{Knowledge Graph:} The decimal equivalent of Drill Size 89 is \hlgreen{0.0091 inches}.
\\
\bottomrule
\end{tabular}
}
\end{table}

Our next experiment examines the ability of LLMs to respond to open-ended questions versus multiple-choice questions. In this section, we seek to understand how the models perform when required to generate answers without predefined options, compared to when they can select from a limited set of potential responses. This approach allows us to assess the effect of external knowledge augmentation in more unconstrained scenarios. Figure \ref{fig:semanticsim} provides the average results across 10 runs for each of the datasets for four different metrics: semantic similarity, ROUGE-1, ROUGE-2, and ROUGE-L. Here, semantic similarity evaluates the closeness in meaning between the generated answer and the reference response by comparing their vector embeddings, thereby capturing the overall semantic content beyond mere word matching. ROUGE-1 measures the overlap of individual words (unigrams) between the generated text and the grounded answer, serving as a baseline indicator of lexical similarity. ROUGE-2 extends this comparison to pairs of consecutive words (bigrams), offering insights into the consistency of short phrases. Finally, ROUGE-L assesses the longest common subsequence between the generated and reference texts, highlighting the structural similarity and the preservation of narrative flow. As shown in Figure \ref{fig:semanticsim}a), for the content specific questions with the additional information from the knowledge graph, the overall scores are higher for most of the models across all metrics. This is especially true for models such as GPT-4o, GPT-4o mini, Gemini 2.0 Flash, Gemini 2.0 Flash-Lite, Llama 3.1 8B Instruct, and Llama 3.2 3B Instruct having an increase of upwards of 3.1$\times$ for ROUGE-1, 2.8$\times$ for ROUGE-2, and 2.9$\times$ for ROGUE-L. These substantial multiplicative gains clearly indicate that the integration of graph-derived information greatly enhances the models’ ability to capture and reproduce relevant ground information thereby significantly improving the quality and fidelity of generated responses in content specific questions. Figure \ref{fig:semanticsim}b) showcases an even more pronounced improvement for open ended generation when testing on the quantitative machining specific queries. For semantic similarity, there is an improvement up to 1.2$\times$ across all models with the exception of Gemini 2.0 Flash-Lite. For ROUGE metrics, all models exhibit an increase in scores compared to baseline with up to 8.5$\times$, 8.8$\times$, and 8.1$\times$ for ROUGE-1, ROGUE-2, and ROUGE-L, respectively. These results provide evidence that more quantitative questions, which demand precise numerical accuracy and technical detail, benefit particularly from the enriched context. While semantic similarity indicates that the basic understanding of quantitative content is already strong, the substantial improvements in ROUGE scores reveal that the additional context helps the models generate far more accurate and detailed quantitative responses.

\begin{table}[ht!]
\centering
\caption{Open ended responses from closed source models Gemini 2.0 Flash, Gemini 2.0 Flash-Lite, GPT-4o, and GPT-4o-mini regarding drill sizes for CNC machining. Grounded quantitative answers are highlighted in \textcolor{yellow}{yellow}, baseline values in \textcolor{red}{red}, and knowledge graph augmented values in \textcolor{green}{green}.}
\label{tab:drill_closed}
\resizebox{\textwidth}{!}{
\begin{tabular}{p{3cm} p{5.33cm} p{5.33cm} p{5.33cm}}
\toprule
\textbf{Model / Question} & \textbf{What is the decimal equivalent of Drill Size 82?} & \textbf{What is the decimal equivalent of Drill Size 84?} & \textbf{What is the decimal equivalent of Drill Size 89?} \\
\midrule
\textbf{Grounded Answer} & 
The decimal equivalent of Drill Size 82 is \hlyellow{0.0125 inches}.
& 
The decimal equivalent of Drill Size 84 is \hlyellow{0.0115 inches}.
& 
The decimal equivalent of Drill Size 89 is \hlyellow{0.0091 inches}. \\
\midrule
\textbf{Gemini 2.0 Flash} & 
\textbf{Baseline:} Drill sizes are often referred to by a number, letter, or fractional inch. Drill size 82 is a numbered drill size. Numbered drill sizes range from size 80, which is \hlred{0.0135 inches}, to size 1, which is \hlred{0.228 inches}. The size increases as the number decreases. Drill size 82 is equal to \hlred{0.0156 inches}.
\newline
\newline
\textbf{Knowledge Graph:} The decimal equivalent of Drill Size 82 is \hlgreen{.0125 inches}.
& 
\textbf{Baseline:} The decimal equivalent of Drill Size 84 is \hlred{0.0095 inches}.
\newline
\newline
\textbf{Knowledge Graph:} The decimal equivalent of Drill Size 84 is \hlgreen{.0115 inches}.
& 
\textbf{Baseline:} The decimal equivalent of Drill Size 89 is \hlred{0.2010 inches}.
\newline
\newline
\textbf{Knowledge Graph:} The decimal equivalent of Drill Size 89 is \hlgreen{.0091 inches}. \\
\midrule
\textbf{Gemini 2.0 Flash-Lite} & 
\textbf{Baseline:} The decimal equivalent of drill size 82 is approximately \hlred{0.0400 inches}.
\newline
\newline
\textbf{Knowledge Graph:} The decimal equivalent of drill size 82 is \hlgreen{.0125 inches}.
& 
\textbf{Baseline:} The decimal equivalent of Drill Size 84 is \hlred{0.040 inches}.
\newline
\newline
\textbf{Knowledge Graph:} The decimal equivalent of drill size 84 is \hlgreen{.0115 inches}.
& 
\textbf{Baseline:} The decimal equivalent of drill size 89 is approximately \hlred{0.040 inches}.
\newline
\newline
\textbf{Knowledge Graph:} The decimal equivalent of drill size 89 is \hlgreen{.0091 inches}. \\
\midrule
\textbf{GPT-4o} & 
\textbf{Baseline:} For drill size 82, which is part of the numbered drill bit system, the decimal equivalent is \hlred{0.0120 inches}.
\newline
\newline
\textbf{Knowledge Graph:} The decimal equivalent of Drill Size 82 is \hlgreen{.0125 inches}.
& 
\textbf{Baseline:} In the standard number drill size system, the smallest size is 80, which corresponds to a diameter of \hlred{0.0135 inches}. If you are referring to a drill size outside of this standard range, it might be a typographical error or a non-standard size specific to a particular manufacturer or application.
\newline
\newline
\textbf{Knowledge Graph:} The decimal equivalent of Drill Size 84 is \hlgreen{.0115 inches}.
& 
\textbf{Baseline:} In the context of standard numbered drill bits, Drill Size 89 is not typically recognized. The standard numbered drill bit sizes go from size 80, which has a decimal equivalent of \hlred{0.0135 inches}.
\newline
\newline
\textbf{Knowledge Graph:} The decimal equivalent of Drill Size 89 is \hlgreen{.0091 inches}. \\
\midrule
\textbf{GPT-4o-mini} & 
\textbf{Baseline:} Drill Size 82 has a diameter of approximately \hlred{0.4130 inches}.
\newline
\newline
\textbf{Knowledge Graph:} The decimal equivalent of Drill Size 82 is \hlgreen{.0125 inches}.
& 
\textbf{Baseline:} Drill size 84 corresponds to a diameter of \hlred{0.0860 inches}.
\newline
\newline
\textbf{Knowledge Graph:} The decimal equivalent of Drill Size 84 is \hlgreen{.0115 inches}.
& 
\textbf{Baseline:} Drill size 89 corresponds to a diameter of approximately \hlred{0.0135 inches}.
\newline
\newline
\textbf{Knowledge Graph:} The decimal equivalent of Drill Size 89 is \hlgreen{.0091 inches}. \\
\bottomrule
\end{tabular}
}
\end{table}

In addition to using quantitative metrics such as semantic similarity and ROUGE scores to evaluate each model's responses, we also include the full answers generated for the quantitative questions. This approach not only highlights the differences between closed source and open source models but also offers deeper insight into how each model integrates the retrieved graph knowledge into its responses. Table \ref{tab:drill_open} and Table \ref{tab:drill_closed} show the baseline and knowledge graph–augmented outputs for drill size questions, which are focused on accurately determining the correct drill size values for open and closed source models, respectively. Correct quantitative answers for each question are highlighted in yellow, baseline LLM values in red, and knowledge graph–augmented values in green for each table. As shown, the use of the knowledge graph consistently improves all open ended outputs of the LLMs' answers by correcting the hallucinations present in the baseline answers. For example, in the open source group, models such as Llama 3.2 3B and Llama 3.1 8B provided baseline answers that deviated significantly from the grounded values (e.g., reporting 0.820 inches or 0.082 inches for drill size 82), whereas, when augmented with the knowledge graph, these models uniformly converged to the correct decimal equivalent of 0.0125 inches. Similarly, Qwen 2.5 7B Instruct confuses the drill size number for its equivalence in inches but this was effectively remedied with information from the graph. The closed sourced models exhibited a similar trend with Gemini 2.0-flash, Gemini 2.0-flash-lite, GPT-4o, and GPT-4o-mini all showing incorrect quantitative answers. These answers varied from nearly correct values, such as GPT-4o's output of 0.012 inches for drill size 82, to very inaccurate ones, including GPT-4o-mini's output of 0.4130 inches for the same drill size. Comparing across open and closed source models, the smaller open sourced models can tend to confuse the drill size number with its corresponding decimal value, effectively mistaking the numeric label for the measurement itself. In contrast, the closed sourced models often produce entirely incorrect values. In some cases, such as with GPT-4o, the model claims that the specified drill size does not exist. These results emphasize the value of incorporating a knowledge graph in open ended scenarios where choices are not given to correctly convert drill sizing information.

\begin{table}[ht]

\centering
\caption{Open ended responses from open source models Llama 3.2 3B Instruct, Llama 3.1 8B Instruct, and Qwen 2.5 7B Instruct regarding parameter selection for CNC machining operation. Grounded quantitative answers are highlighted in \textcolor{yellow}{yellow}, baseline values in \textcolor{red}{red}, and knowledge graph augmented values in \textcolor{green}{green}.}
\label{tab:operation_open}
\resizebox{\textwidth}{!}{
\begin{tabular}{p{3cm} p{5.33cm} p{5.33cm} p{5.33cm}}
\toprule
\textbf{Model / Question} & \textbf{For drilling operations on steel (4140), what cutting speed (SFM) is recommended?} & \textbf{For tapping operations on aluminum, what cutting speed (SFM) is recommended?} & \textbf{For milling operations on brass with a tool diameter between 0.5 and 1.0 inches, what chip load is recommended?}\\
\midrule
\textbf{Grounded Answer} & 
For drilling operations on steel (4140), a cutting speed of \hlyellow{90 SFM} is recommended.
& 
For tapping operations on aluminum, a cutting speed of \hlyellow{100 SFM} is recommended. & For milling operations on brass with a tool diameter between 0.5 and 1.0 inches, a chip load of \hlyellow{0.004 inches per revolution} is recommended.\\
\midrule
\textbf{Llama 3.1 8B Instruct} & 
\textbf{Baseline:} The recommended cutting speed for drilling operations on 4140 steel ranges from \hlred{50 to 300 SFM}, depending on the drill bit type and desired finish. 
\newline
\newline
\textbf{Knowledge Graph:} For drilling operations on steel (4140), the recommended cutting speed is \hlgreen{90 sfm}.
& 
\textbf{Baseline:} In general, the recommended cutting speed for tapping aluminum ranges from \hlred{50 to 150 SFM}, depending on the specific conditions.
\newline
\newline
\textbf{Knowledge Graph:} For tapping operations on aluminum, the recommended cutting speed is \hlgreen{100 sfm}.
&
\textbf{Baseline:} A commonly cited range for chip loads on brass is between \hlred{0.002 and 0.008 inches per tooth (ipt)}.
\newline
\newline
\textbf{Knowledge Graph:} For milling operations on brass with a tool diameter between 0.5 and 1.0 inches, the recommended chip load is \hlgreen{0.004 ipr}. 
\\
\midrule
\textbf{Llama 3.2 3B Instruct} & 
\textbf{Baseline:} For drilling operations on 4140 steel, the recommended cutting speeds typically range from \hlred{60 to 120 SFM}, with the optimal speed depending on the specific drilling conditions.
\newline
\newline
\textbf{Knowledge Graph:} For drilling operations on steel (4140), a cutting speed of \hlgreen{90 SFM} is recommended.
& 
\textbf{Baseline:} A cutting speed of \hlred{30 to 50 FPM} is commonly recommended for tapping operations on aluminum alloys, but the optimal cutting speed may vary depending on the application, tooling, and machining conditions.
\newline
\newline
\textbf{Knowledge Graph:} For tapping operations on aluminum, a recommended cutting speed is \hlgreen{100 sfm}.
&
\textbf{Baseline:} For milling operations on brass, a general rule of thumb is to use a chip load that is between \hlred{0.002 and 0.005 inches} for a tool diameter between 0.5 and 1.0 inches.
\newline
\newline
\textbf{Knowledge Graph:} Based on the provided context, for milling operations on brass with a tool diameter between 0.5 and 1.0 inches, the recommended chip load is \hlgreen{0.004 inches per revolution (ipr)}.
\\
\midrule
\textbf{Qwen 2.5 7B Instruct} & 
\textbf{Baseline:} The recommended SFM for drilling 4140 steel generally ranges between \hlred{25 and 35 feet per minute (FPM)}.
\newline
\newline
\textbf{Knowledge Graph:} For drilling operations on steel (4140), a recommended cutting speed (SFM) is \hlgreen{90 sfm}. 
& 
\textbf{Baseline:} For tapping operations on aluminum, a recommended cutting speed (Surface Footage per Minute, SFM) typically ranges between \hlred{100 to 200 SFM}.
\newline
\newline
\textbf{Knowledge Graph:} For tapping operations on aluminum, a recommended cutting speed (SFM) is \hlgreen{100 sfm}. 
&
\textbf{Baseline:} A common starting point for chip loads in this range is approximately \hlred{0.003 to 0.005 inches per tooth (IPM)}.
\newline
\newline
\textbf{Knowledge Graph:} For milling operations on brass with a tool diameter between 0.5 and 1.0 inches, a chip load of \hlgreen{0.004 inches per revolution (ipr)} is recommended. 
\\
\bottomrule
\end{tabular}
}
\end{table}

\begin{table}[h!]
\centering
\caption{Open ended responses from closed source models Gemini 2.0 Flash, Gemini 2.0 Flash-Lite, GPT-4o, and GPT-4o-mini regarding parameter selection for CNC machining operation. Grounded quantitative answers are highlighted in \textcolor{yellow}{yellow}, baseline values in \textcolor{red}{red}, and knowledge graph augmented values in \textcolor{green}{green}.}
\label{tab:operation_closed}
\resizebox{\textwidth}{!}{
\begin{tabular}{p{3cm} p{5.33cm} p{5.33cm} p{5.33cm}}
\toprule
\textbf{Model / Question} & \textbf{For drilling operations on steel (4140), what cutting speed (SFM) is recommended?} & \textbf{For tapping operations on aluminum, what cutting speed (SFM) is recommended?} & \textbf{For milling operations on brass with a tool diameter between 0.5 and 1.0 inches, what chip load is recommended?}\\
\midrule
\textbf{Grounded Answer} & 
For drilling operations on steel (4140), a cutting speed of \hlyellow{90 SFM} is recommended.
& 
For tapping operations on aluminum, a cutting speed of \hlyellow{100 SFM} is recommended. & For milling operations on brass with a tool diameter between 0.5 and 1.0 inches, a chip load of \hlyellow{0.004 inches per revolution} is recommended.\\
\midrule
\textbf{Gemini 2.0-flash} & 
\textbf{Baseline:} For 4140 steel, which is a medium-carbon, low-alloy steel known for its high strength and good hardenability, typical cutting speeds for drilling operations generally range from \hlred{70 to 120 SFM}.
\newline
\newline
\textbf{Knowledge Graph:} For drilling operations on steel (4140), a cutting speed of \hlgreen{90 SFM} is recommended.
& 
\textbf{Baseline:} For general-purpose tapping of aluminum alloys like 6061-T6, a common starting point is between \hlred{30 to 60 SFM}. However, softer alloys like 1100 or 3003 can often be tapped at higher speeds, potentially up to \hlred{75 SFM} or more.
\newline
\newline
\textbf{Knowledge Graph:} For tapping operations on aluminum, a cutting speed of \hlgreen{100 SFM} is recommended.
&
\textbf{Baseline}: The recommended chip load for milling brass with a tool diameter between 0.5 and 1.0 inches typically falls within the range of \hlred{0.001 to 0.004 inches} per tooth.
\newline
\newline
\textbf{Knowledge Graph:}  For milling operations on brass with a tool diameter between 0.5 and 1.0 inches, a chip load of \hlgreen{0.004 inches per revolution (ipr)} is recommended.
\\
\midrule
\textbf{Gemini 2.0-flash-lite} & 
\textbf{Baseline:} For high-speed steel (HSS) drills, a starting point for the cutting speed when drilling 4140 steel is typically in the range of \hlred{40 to 60 surface feet per minute} (SFM) for the steel in a hardened state.
\newline
\newline
\textbf{Knowledge Graph:} For drilling operations on steel (4140), the recommended cutting speed is \hlgreen{90 SFM}.
& 
\textbf{Baseline:}  For most aluminum alloys, a starting point for the cutting speed is typically in the range of \hlred{50 to 200 surface feet per minute (SFM)}. 
\newline
\newline
\textbf{Knowledge Graph:} For tapping operations on aluminum, a cutting speed of \hlgreen{100 SFM (Surface Feet per Minute)} is recommended.
&
\textbf{Baseline:} For milling brass with a tool diameter between 0.5 and 1.0 inches, a chip load of \hlred{0.003 to 0.010 IPT} is a good starting point.
\newline
\newline
\textbf{Knowledge Graph:} For milling operations on brass with a tool diameter between 0.5 and 1.0 inches, a chip load of \hlgreen{0.004 inches per revolution (ipr)} is recommended.
\\
\midrule
\textbf{GPT-4o} & 
\textbf{Baseline:} For annealed AISI 4140 steel, which typically has a hardness of around 150-200 Brinell, the recommended cutting speed is usually in the range of \hlred{70 to 100 surface feet per minute (SFM)} when using high-speed steel (HSS) drill bits.
\newline
\newline
\textbf{Knowledge Graph:} For drilling operations on steel (4140), a cutting speed of \hlgreen{90 SFM} is recommended.
& 
\textbf{Baseline:} The recommended cutting speed for tapping aluminum typically ranges from \hlred{75 to 150 surface feet per minute (SFM)}.
\newline
\newline
\textbf{Knowledge Graph:} For tapping operations on aluminum, a cutting speed of \hlgreen{100 surface feet per minute (SFM)} is recommended.
&\textbf{Baseline:} For general milling operations on brass with a tool diameter between 0.5 and 1.0 inches, a typical chip load recommendation would range from approximately \hlred{0.002 to 0.010 inches per tooth (IPT)}.
\newline
\newline
\textbf{Knowledge Graph:} For milling operations on brass with a tool diameter between 0.5 and 1.0 inches, the recommended chip load is \hlgreen{0.004 inches per revolution (ipr)}.
\\
\midrule
\textbf{GPT-4o-mini} & 
\textbf{Baseline:} Typically, for drilling operations on 4140 steel, a cutting speed in the range of \hlred{60 to 100 SFM} is often recommended when using high-speed steel (HSS) drill bits.
\newline
\newline
\textbf{Knowledge Graph:} For drilling operations on steel (4140), the recommended cutting speed is \hlgreen{90 surface feet per minute (SFM)}.
& 
\textbf{Baseline:} For most aluminum alloys, a common recommendation for tapping speeds is in the range of \hlred{100 to 300 SFM}.
\newline
\newline
\textbf{Knowledge Graph:} For tapping operations on aluminum, the recommended cutting speed is \hlgreen{100 surface feet per minute (SFM)}.
& {Baseline:} For brass, which is a relatively soft and ductile material, the recommended chip load generally falls within the range of \hlred{0.002 to 0.008 inches per tooth}.
\newline
\newline
\textbf{Knowledge Graph:} For milling operations on brass with a tool diameter between 0.5 and 1.0 inches, the recommended chip load is \hlgreen{0.004 inches per revolution (ipr)}.\\
\bottomrule
\end{tabular}
}
\end{table}

Table \ref{tab:operation_open} and Table \ref{tab:operation_closed} present open and closed sourced model responses, respectively, to quantitative questions regarding selection of correct process parameters, which is especially important for ensuring optimal machining performance, precise material removal, and enhanced tool longevity across varying operational conditions. Similar to the previously mentioned tables, correct quantitative answers for each question are highlighted in yellow, baseline LLM values in red, and knowledge graph–augmented values in green for each table. For the open sourced models, the baseline outputs typically provide a broad range of values such as Llama 3.2 3B Instruct suggesting a range from 60 to 120 surface feet per minute (SFM), and Llama 3.1 8B Instruct suggesting a range of 50 to 300 SFM. These wide ranges reflect uncertainty and a reliance on generic guidelines which do not allow a user to determine the optimal process parameters specified in technical documents. However, when the LLMs are augmented with knowledge graph information, these models converge on a precise recommendation of 90 SFM for drilling 4140 steel, 100 SFM for tapping aluminum, and a chip load of 0.004 inches per revolution (IPR) for milling brass, correctly aligning with the grounded answers. This shift from broad ambiguous ranges to a single well-defined value highlights the corrective influence of domain-specific information. The closed source LLMs exhibit a similar trend, frequently providing parameter recommendations as broad ranges that mirror inherent uncertainty and a lack of industry-specific technical standards. However, when augmented with the knowledge graph the models once again converge to the correctly recommended process parameters. Comparing across open and closed models reveals that all tend to recommend broad and imprecise parameter ranges for operations such as drilling, tapping, and milling. Regardless of model size or architecture, the baseline outputs consistently display wide ranges that fail to account for material properties and precise machining requirements. This suggests that the inherent complexity and scale of the models do not necessarily translate into accurate internal representations of domain-specific machining standards. In summary, without integrating knowledge graph information the outputs remain too generic to guide optimal parameter selection demonstrating that the incorporation of structured, domain-specific data is essential for constraining outputs and calibrating process parameters to align with validated manufacturing standards.

\section{Conclusion}\label{conclusion}
This paper introduces ARKNESS (Augmented Retrieval Knowledge Network Enhanced Search and Synthesis), a hybrid LLM-agnostic pipeline that couples a semantically rich automatically constructed knowledge graph with any large language model to provide answers that are grounded and numerically precise. ARKNESS provides traceable answers validated in technical information through paragraph level entity-relation extraction that converts heterogeneous document formats into triples while retaining their original contextual information. Then, by storing the knowledge graph in a relational database, we can leverage vector‑embedding search alongside beam search to retrieve the most relevant triples and contextual information, thereby conditioning the LLM to deliver precise, technically validated answers to user queries. Experiments on multiple choice questions across content specific and machining specific queries showcases uplifts up to 25\% in accuracy and 22.4\% in F1-score. For opened ended questions, uplifts up to 8.1$\times$ for ROGUE-L scores demonstrate that the framework improves the structural and semantic information of generated responses relative to the grounded answers. These advantages allow the elimination of hallucinations prevalent in LLMs thereby enhancing process planning and usefulness in CNC machining contexts. Future work will focus on extending ARKNESS to a heterogeneous multimodal knowledge graphs, further strengthening factual grounding to enable closed-loop decision support in advanced manufacturing enviornments.


\section*{Declaration of Competing Interest}
The authors declare that they have no known competing financial interests or personal relationships that could have appeared to influence the work reported in this paper.

\section*{Acknowledgments}
This work was supported under Cooperative Agreement W56HZV-21-2-0001 with the US Army DEVCOM Ground Vehicle Systems Center (GVSC), through the Virtual Prototyping of Autonomy Enabled Ground Systems (VIPR-GS) program.

DISTRIBUTION STATEMENT A. Approved for public release; distribution is unlimited. OPSEC9768

 \bibliographystyle{elsarticle-num-names} 
 \bibliography{references}


\end{document}